\documentclass{easychair}

\usepackage{doc}
\usepackage[]{algorithm2e} %for algorithm
\usepackage{adjustbox}
\usepackage{graphicx}
\usepackage{float}
\usepackage{amsmath,amsfonts,amsthm}
\usepackage{tikz}
\usepackage{hyperref}

\usepackage{enumitem}
\setlist[description]{leftmargin=\parindent,labelindent=\parindent}

\DeclareMathOperator*{\argmax}{argmax}
\DeclareMathOperator*{\argmin}{argmin}

\title{On the Robustness of Active Learning}

\author{
Lukas Hahn\inst{1}\textsuperscript{,}\inst{2}
\and
   Lutz Roese-Koerner\inst{2}
\and
   Peet Cremer\inst{2}
\and
   Urs Zimmermann\inst{2}
\and
   \\Ori Maoz\inst{2}
\and
   Anton Kummert\inst{1}   
}

\institute{
  University of Wuppertal,\\
  Department of Electrical Engineering\\
  Wuppertal, Germany\\
  \email{lukas.hahn@uni-wuppertal.de}
\and
   Aptiv,\\
   Wuppertal, Germany\\
   \email{firstname.lastname@aptiv.com}\\
 }

\authorrunning{Hahn, Roese-Koerner, Cremer, Zimmermann, Maoz and Kummert}

\titlerunning{On the Robustness of Active Learning}

\newcommand\copyrighttext{%
	\centering
	\small As published in the proceedings of the 5th Global Conference on Artificial Intelligence (GCAI),\\ EPiC Series in Computing, Volume 65, pages 152-162, \href{https://doi.org/10.29007/thws}{DOI: 10.29007/thws}, 2019.}
\newcommand\copyrightnotice{%
	\begin{tikzpicture}[remember picture,overlay]
	\node[anchor=south,yshift=30pt] at (current page.south) {\fbox{\parbox{\dimexpr\textwidth-\fboxsep-\fboxrule\relax}{\copyrighttext}}};
	\end{tikzpicture}%
}

\begin{document}

\maketitle

\begin{abstract}
Active Learning is concerned with the question of how to identify the most useful samples for a Machine Learning algorithm to be trained with. When applied correctly, it can be a very powerful tool to counteract the immense data requirements of Artificial Neural Networks. However, we find that it is often applied with not enough care and domain knowledge. As a consequence, unrealistic hopes are raised and transfer of the experimental results from one dataset to another becomes unnecessarily hard.

In this work we analyse the robustness of different Active Learning methods with respect to classifier capacity, exchangeability and type, as well as hyperparameters and falsely labelled data. Experiments reveal possible biases towards the architecture used for sample selection, resulting in suboptimal performance for other classifiers.
We further propose the new "Sum of Squared Logits" method based on the Simpson diversity index and investigate the effect of using the confusion matrix for balancing in sample selection. 

\end{abstract}

\copyrightnotice

% The table of contents below is added for your convenience. Please do not use
% the table of contents if you are preparing your paper for publication in the
% EPiC Series or Kalpa Publications series

%\setcounter{tocdepth}{2}
%{\small
%\tableofcontents}

%\section{To mention}
%
%Processing in EasyChair - number of pages.
%
%Examples of how EasyChair processes papers. Caveats (replacement of EC
%class, errors).

\section{Introduction}
\label{sec:intro}
The term Active Learning describes the field of selecting samples from a given pool of data in order to subsequently train a Machine Learning algorithm with. This can be done for two major reasons: Firstly, deciding which subset of collected data will be annotated in order to create training and validation set for a supervised Machine Learning task. While it can be comparatively easy and inexpensive to record and gather sensor data and ever decreasing cost makes it affordable to possibly neglect storage expenses, reliable ground truth annotation still requires manual labour and is therefore the crucial factor. In industrial application of Machine Learning to various tasks, budget and time constraints play a significant role and performance can depend on choosing the best $n$ samples to train on.

Secondly, one could think of using Active Learning methods as a form of regularization. While increasing the number of available training samples is in general regarded as helpful, certain factors can lead to an impaired performance when doing so. The more objects in a recognition task are standardized, the more redundant information is potentially added to the dataset with each new sample, which can result in worsened generalization. Active Learning methods can also be applied to sanitize a dataset from falsely labelled samples, as a suitable strategy will not pick samples with a conspicuous difference between label and prediction.

However, despite the great potential of Active Learning it also bears significant risks. If applied in an incorrect way it could lead to a sub-optimal sample selection, and, in the worst case, rendering the complete Machine Learning task unsuccessful. In order to point out how to avoid these pitfalls, we examine a set of known Active Learning query strategies, as well as some extensions of our own, and their performance on various different image classification datasets. We then view their performance under different aspects including changing hyperparameters, influence of falsely labelled data and the replaceability of varying CNN architectures. Eventually we regard the performance of the same strategies when applied to a problem of hierarchical classifiers. Our main contributions are:\\
1.) A robustness investigation of state-of-the-art Active Learning strategies with respect to the impact of falsely labelled data, hyperparameters and the impact of changing the classifier model during the selection phase. 2.) An extension of the Active Learning method based on Entropy computation using the Simpson Diversity. 3.) Theoretical insights and experimental results for Active Learning on Hierarchical Neural Networks.

%\begin{itemize}
%  \item A robustness investigation of State-of-the-art Active Learning strategies with respect to the impact of falsely labelled data, hyperparameters and the impact of changing the classifier model during the selection phase.
%  \item An extension of the Active Learning method based on Entropy computation using the Simpson Diversity.
%  \item Theoretical insights and experimental results for Active Learning on Hierarchical Neural Networks.
%\end{itemize}

%\subsection{Citations}
%When citing a multi-author paper, use \etal \cite{Authors06}

%------------------------------------------------------------------------
\section{Related Work}
\label{sec:relWork}
%Active Learning has been a vivid research area for more than two decades. With the rise of Deep Learning and growing data requirements, it is now possibly more important than ever. %The following paragraphs should provide the reader with a rough overview of the topic, but by no means claim to be complete.

% \subsection{Pre Deep Learning}
% \label{subsec:preDL}
% Pre deep Learning
An overview of methods from the pre Deep Learning area can be found in the very comprehensive review of \cite{settles_10}. Many approaches originating from that time (e.g. Uncertainty sampling, Margin based sampling, Entropy Sampling, ...) have been later adapted to neural networks. Additional examples for this include the approach of \cite{roy-mccallum_01}, who applied a Monte Carlo method to compute an estimated error reduction that can be used for sample selection as well as clustering approaches like those described in \cite{nguyen-smeulders_04} and \cite{dasgupta-hsu_08}. 

%Furthermore, methods from the field of instance selection (c.f. \cite{olvera-lopez-martinez-trinidad_10}) are closely related, as they treat a very similar problem. The only difference here is that most of the labels in Active Learning are assumed to be unknown, which is not the case for instance selection.

% \subsection{Unsupervised / Semi-supervised Learning}
% \label{subsec:unsup}
% Unsupervised
%Most Active Learning Methods are applied for tasks of supervised learning. However, there are some exceptions:
%\cite{chalapathy-etal_18} for example suggest an anomaly detection in a completely unsupervised setting. The authors do not call this Active Learning, but the underlying idea shows similarities. 
%Semi supervised
\cite{wang-etal_17} and \cite{rottmann-etal_18} propose a semi-supervised approach. They use Active Learning to query samples which the network has not yet understood and use label propagation to also utilize well understood samples with "pseudo-labels".

% \subsection{Supervised Learning}
% \label{subsec:sup}
In the field of supervised learning, \cite{korattikara-etal_15} used a Bayes approach to distil a Monte Carlo approximation of the posterior predictive density for sample selection. In the theoretical work of \cite{kabkab-etal_16}, Active Learning was rephrased as a convex optimisation problem and the balancing of the selection of samples with high diversity and those that are very representative for a subset are discussed.
Unlike many other methods, the core-set approach of \cite{sener-savarese_18} does not use the output layer of a network for Active Learning. Instead they solve a relaxed $k$-centres problem to minimize the maximal distance to the closest cluster centre for each sample in a space that is spanned by the neurons of a hidden layer of a network. As discussed later, this approach has a very high independence of the actual classes of a network, which can be helpful when dealing with hierarchical networks \cite{weyers-etal_18} for example.

\cite{gal-etal_17} introduced the concept of live-dropout to Active Learning. The idea is to approximate the behaviour of an ensemble of Bayesian estimators by activating dropout during inference and multiple forward passes. They furthermore developed an Active Learning framework which is able to use this and other deep Bayesian methods. In the same line of thought, \cite{ducoffe-precioso_17} investigated live dropout and Query-by-committee methods.
However, \cite{beluch-etal_18} used ensembles of CNNs with identical architectures but different weight initiations to show that ensembles work better than "ensemble approximation methods" like the above mentioned MC dropout of \cite{gal-etal_17} or approaches based on geometric distributions like \cite{sener-savarese_18}. 

%\subsection{Meta learners}
Some recent approaches also utilize "meta" knowledge for Active Learning. \cite{fang-etal_17} introduced "Policy based Active Learning". There, reinforcement learning is used for stream based Active Learning in a language processing setting. This is very similar to the approach of \cite{bachman-etal_17} who proposed "Learning Algorithms for Active Learning". They also used Reinforcement Learning to jointly learn a data representation, an item selection heuristic and a method for constructing prediction functions from labelled training sets.
\cite{heilbron-etal_18} reuse knowledge from previously annotated datasets to improve the Active Learning performance.

% LiDAR
% Furthermore, research with multi-sensor approaches is performed. \cite{feng-etal_19} use a combination of 2D RGB image detectors, 3D LiDAR point cloud classifiers, Monte Carlo Dropout and Deep Ensemble classifiers for Active Learning on the KITTI dataset.

%------------------------------------------------------------------------
\section{Methods}
\label{methods}

In the following we review existing methods from the field of pool-based Active Learning and propose a suggestion of our own. Given a classification model $\theta$ and a dataset $\mathcal{D}$, consisting of a feature and label pair $\langle x\in X, y\in Y \rangle$, such an algorithm has the following structure:
\begin{algorithm}[H]

	\SetKwData{Left}{left}\SetKwData{This}{this}\SetKwData{Up}{up}
	\SetKwFunction{Union}{Union}\SetKwFunction{FindCompress}{FindCompress}
	\SetKwInOut{Input}{Input}\SetKwInOut{Output}{Output}

	\Input{$\mathcal{L} \subset \mathcal{D}  = \lbrace\langle x, y \rangle\rbrace$ : Labelled set,\\
	$\mathcal{U} = \mathcal{D}\, \backslash \, \mathcal{L}  = \lbrace\langle x, ? \rangle\rbrace$ : Unlabelled set, \\
	$\theta$ : Classification model, \\
	$\phi$: Active Learning query function \\ }

	\While{$\vert\mathcal{U} \vert \, > 0 \; \wedge $ no stopping criterion}{
		$\theta$ = $train(\mathcal{L})$\;
	\For{\bf{all} $\langle x, ?\rangle \in \mathcal{U}$}{
		Compute Active Learning metric $\phi(x)$ under $\theta$\;
		}
		Choose $x^\star$ with highest magnitude of $\phi(x^\star)$ \;
		Annotate $y^\star$\;
		$\mathcal{L} \leftarrow \mathcal{L} \cup \lbrace \langle x^\star, y^\star\rangle \rbrace$\;
		$\mathcal{U} \leftarrow \mathcal{U} \setminus \lbrace \langle x^\star, y^\star \rangle \rbrace$\;
	}

\caption{Pool-Based Active Learning.}
\end{algorithm}
Considering a large dataset, one can query numerous samples at once. This set of the chosen samples is denoted by $\mathcal{B} \subset U$ \cite{kabkab-etal_16}. 

We take a closer look at uncertainty sampling, a strategy that selects samples the classifier is uncertain about. In this context uncertainty means a low confidence for the predicted class that is given by $\hat{y} = \argmax_y \, P_\theta(y|x)$. We consider three commonly used uncertainty measures: 
%\begin{center}
\begin{enumerate}
	\item[(a)] \textit{Least Confident:} $x^\star_{LC} = \argmax_x \, (1 - P_\theta(\hat{y} |x))$ %$x^\star_{LC} = \underset{x}{\argmax} \, (1 - P_\theta(\hat{y} |x))$
	\item[(b)] \textit{Margin:} $x^\star_M = \argmin_x \, (P_\theta(\hat{y}_1|x) - P_\theta(\hat{y}_2|x))$
	\item[(c)] \textit{Entropy:} $x^\star_H = \argmax_x \left( - \sum_{i}\, P_\theta(y_i | x) \log P_\theta(y_i | x)\right)$
\end{enumerate}
%\end{center}
(a): Considering only one class label, the sample $x^\star_{LC}$ with the least confident label prediction is selected. (b): Margin sampling includes information about the second most certain prediction. The algorithm queries the sample $x^\star_M $ with the smallest difference between the two most probable class labels. (c): For multi class tasks, it is relevant to consider all label confidences. For each sample every class probability is weighted with its information content and summed up. The algorithm queries the sample with the highest entropy $x^\star_H $ \cite{settles_10}.

For the following experiments we implement eight query strategies.\\
Based on Least Confident (a):
\vspace{-0.2cm}
\begin{description}\itemsep-3pt
	\item[Naive Certainty (NC) Low:] Select $n$ samples with the minimal maximal activation in classifier logits. Since basing the decision only on the one highest activated neuron is a very straightforward approach, we call this family of strategies the "Naive" methods.
	\item[NC Range:] Select $n$ samples within a certain range of the classifier logits' activation (e.g. $[0.1, 0.9]$).
	\item[NC Diversity:] Select $n$ samples with the minimal maximal activation in classifier logits and additionally prevent that similar samples are chosen by calculating the diversity of the samples below the threshold compared to those already included in the training set.
	\item[NC Balanced:] Select $n$ samples with the minimal maximal activation in classifier logits and balance the class distribution using the reciprocal value of the classification confusion matrix obtained with the previous training set. Terminates if one class contains no more samples to be drawn.
\end{description}
Based on Margin (b):
\vspace{-0.2cm}
\begin{description}\itemsep-3pt
	\item[Margin:] Select $n$ samples with the smallest difference of the two highest firing logits.	
\end{description}
Based on Entropy (c):
\vspace{-0.2cm}
\begin{description}\itemsep-3pt
	\item[Entropy High:] Select $n$ samples with the highest entropy.
	\item[Sum of Squared Logits (SOSL):] Select $n$ samples with the highest Simpson diversity index $D = 1 - \sum_i (l_i)^2$ \cite{simpson} (cf. \ref{subsec:SOSL}).
\end{description}
%\begin{description}\itemsep-3pt
\textbf{Core Set Greedy:} A similarity measure in the embedding space. Creates a core set by approximating the %metric $k$-centre
problem of distributing $k$-centres in $n$ points, such that the minimal distance of all points to the nearest centre is maximized. Select $n$ samples for which the minimum distance to all samples which are already part of the training set is maximized (cf. \cite{sener-savarese_18}).
%\end{description}

\subsection{Sum of Squared Logits (SOSL) Method}
\label{subsec:SOSL}
In Active Learning, we require a measure of how sure the classifier is that its class decision during inference is accurate. One possibility for such an accuracy-of-inference measure is to analyze the distribution of logits. Within the trained model of the classifier, the logits can be interpreted as probabilities that the inferred sample belongs to the class associated of the respective logit. If the logits are strongly biased in favour of a certain class, it is very likely that the given sample belongs to the class corresponding to the strongest logit. On the contrary, if the logits do not show a clear preference for a certain class, there is a high risk that taking the class of the strongest logit results in a false prediction. In other words, to which degree the distribution of logits tends towards peaks rather than an equipartition indicates how accurate the inference is going to be.

In previous literature, the Shannon entropy \cite{shannon} has been frequently used as a measure of how peaked or equipartitioned a distribution is. A valid strategy for Active Learning could then be to sort out those samples, for which the Shannon entropy $H = - \sum_i l_i \log(l_i)$, with $l_i$ being the values of the logits, is particularly high. However, a shortcoming of this approach is that it does not adequately account for the situation when the the distribution of logits is admittedly strongly peaked, but with peaks on more than one class logit. Such a situation can easily arise in samples, when they belong to classes showing similarities and the classifier's model does not yet feature a clear decision boundary between them. In such a case, the distribution of logits is still far away from an equipartition, resulting in a relatively low value for the Shannon entropy $H$. Thus, although labelling these samples would be particularly valuable for fleshing out the decision boundary and allowing the classifier to better separate between classes, they would not be added to Active Learning training set.

To overcome these shortcomings of the Shannon entropy $H$ as a measure for characterizing the distribution of logits $l_i$, we propose to use the Simpson diversity index $D = 1 - \sum_i (l_i)^2$ \cite{simpson} instead. The closer the distribution $l_i$ is to an equipartition, the larger $D$ becomes. If the $l_i$ shows a strong peak at a certain $i$, $D$ is close to zero. Finally, if the $l_i$ are strongly peaked among several classes, $D$ will have a small-to-moderate value between zero and one. The latter property of $D$ in particular allows to select those samples for labelling, for which the classifier can narrow the class decision down to a few classes, among which it is still unsure. The Active Learning strategy is then to select in each iteration the $n$ samples with highest $D$.

%------------------------------------------------------------------------
\section{Experiments and Results}

We conduct a series of experiments with the query strategies presented in section \ref{methods} on six different datasets for image classification (cf. Table \ref{tab:datasets}). These consist of the well-known digit classification set MNIST \cite{mnist} and the thereof inspired dataset of the Latin alphabet CoMNIST \cite{comnist} and clothing classification Fashion-MNIST \cite{fmnist}, as well as general object classification CIFAR-10 \cite{cifar} and the house number collection SVHN \cite{svhn}. We furthermore evaluate strategies on a private dataset of $33$ different classes of traffic signs (TSR) represented through small grey scale images.
\begin{table}[h!]
	\centering %\begin{center} % \centering bringt die caption näher an's table
	\begin{adjustbox}{max width=\textwidth}
	\begin{tabular}{l|cccccc}
								& CIFAR-10		& CoMNIST 		& Fashion-MNIST	& MNIST 		& SVHN			& TSR   			\\ \hline \hline
		Classes            		& $10$			& $26$      	& $10$			& $10$    		& $10$			& $33$  			\\ 
		Image Size         		& $32\times32$	& $32\times32$ 	& $28\times28$	& $28\times28$ 	& $32\times32$	& $34\times34$   	\\
		Channels			 	& $3$			& $1$ 			& $1$			& $1$ 			& $3$			& $1$ 				\\
		Training Samples   		& $50\,000$		& $9\,918$    	& $60\,000$		& $60\,000$ 	& $73\,257$		& $265\,774$	  	\\
		Validation Samples 		& $10\,000$ 	& $1\,300$    	& $10\,000$ 	& $10\,000$ 	& $26\,032$ 	& $66\,443$			\\
		
	\end{tabular}
	\end{adjustbox}
	%\end{center}
	\caption{Characteristics of the datasets used for the experiments.}
	\label{tab:datasets}
\end{table}
\begin{figure}[H]%[h!]
	\begin{center}
		\begin{tabular}{cr@{}cr@{}}%{@{}c@{\hspace{.5cm}}c@{}}
       		\includegraphics[clip, trim=0.5cm 0.65cm 1.65cm 1.4cm,
       			page=1,width=.375\textwidth]{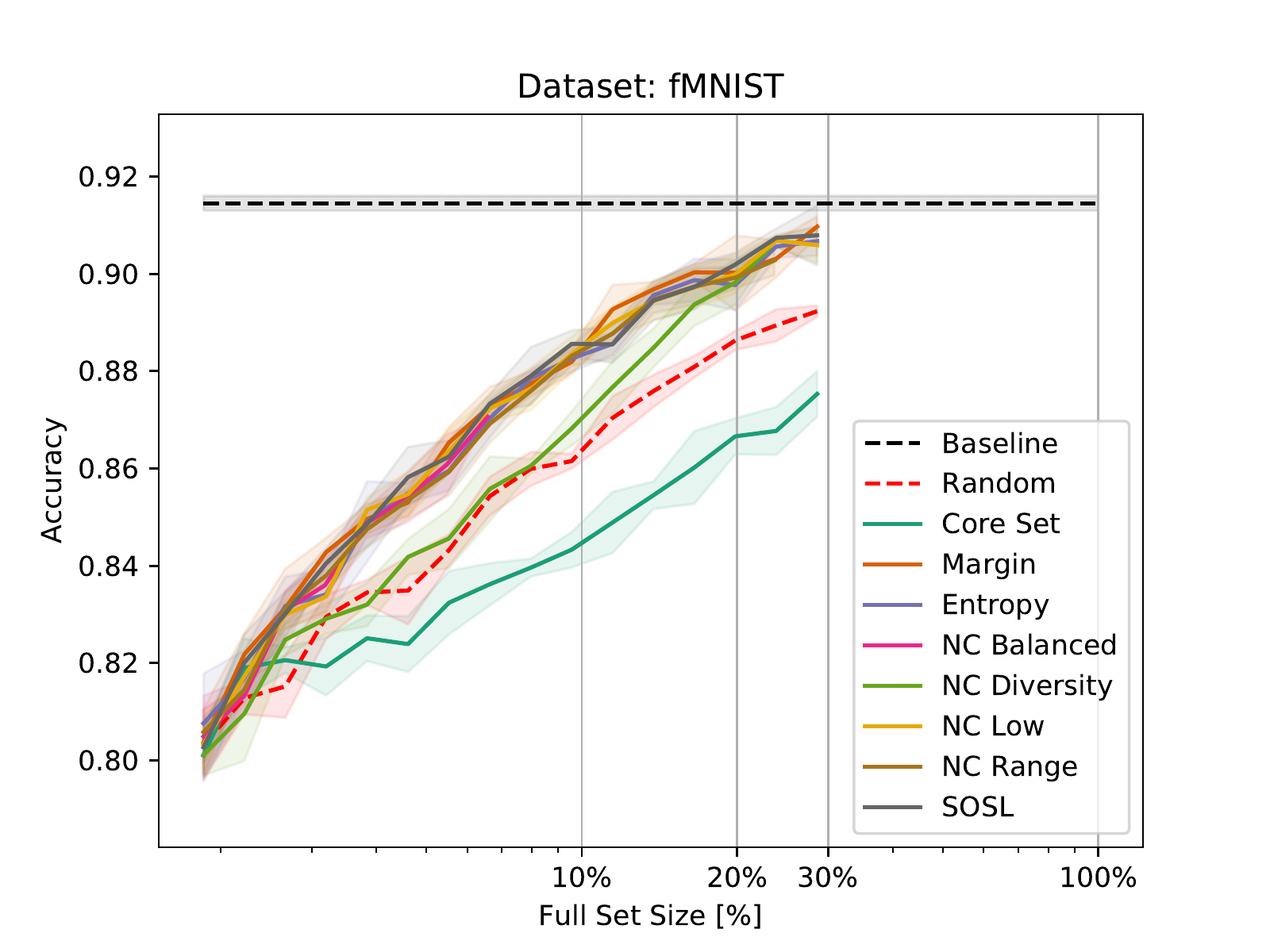} & \hspace{0.0cm}
       		\includegraphics[clip, trim=0.5cm 0.65cm 1.65cm 1.4cm,
       			page=1,width=.375\textwidth]{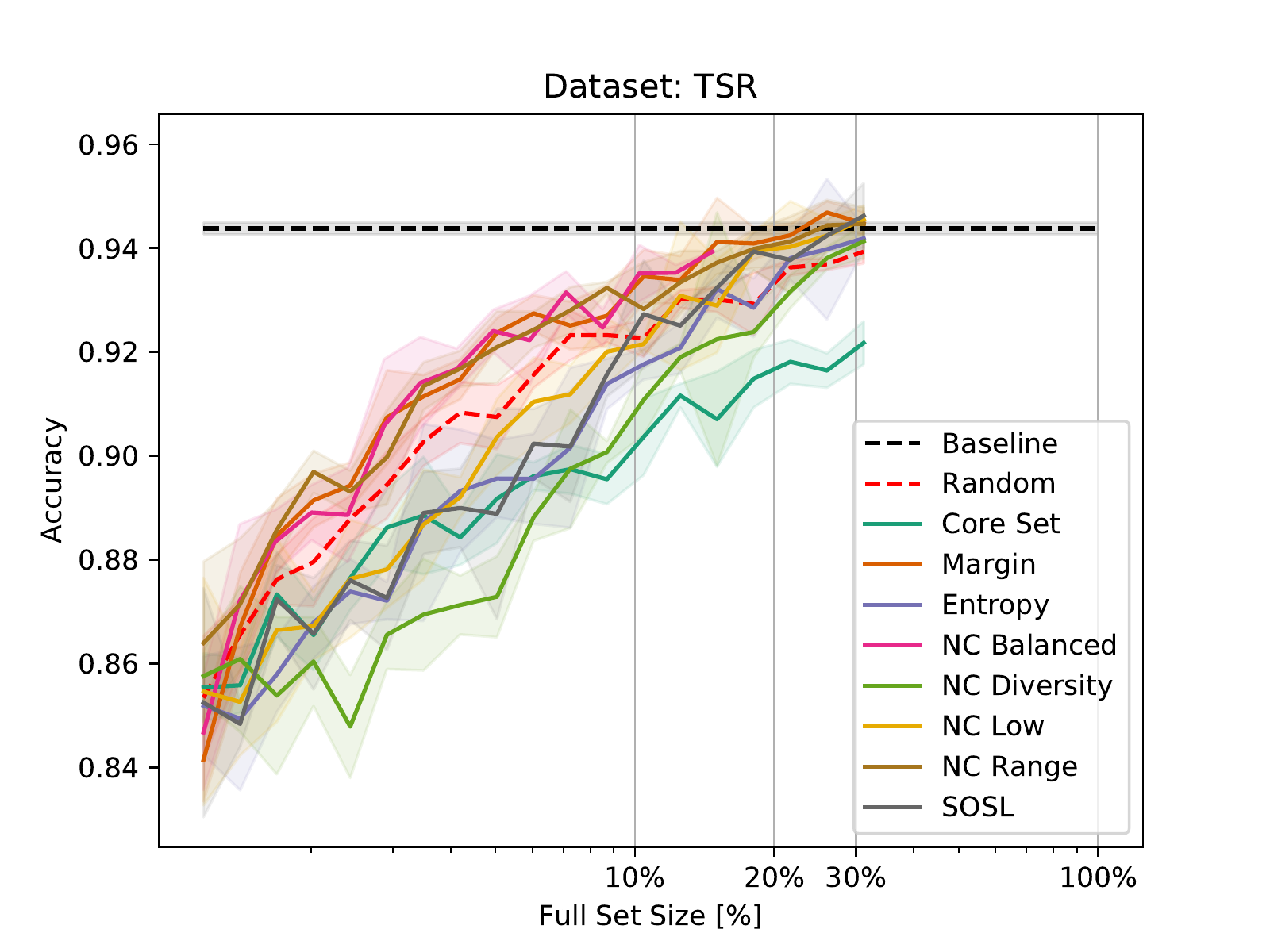} \\ %& \hspace{-0.55cm}
       		
	     	\includegraphics[clip, trim=0.5cm 0.35cm 1.65cm 1.4cm,
	     		page=1,width=.375\textwidth]{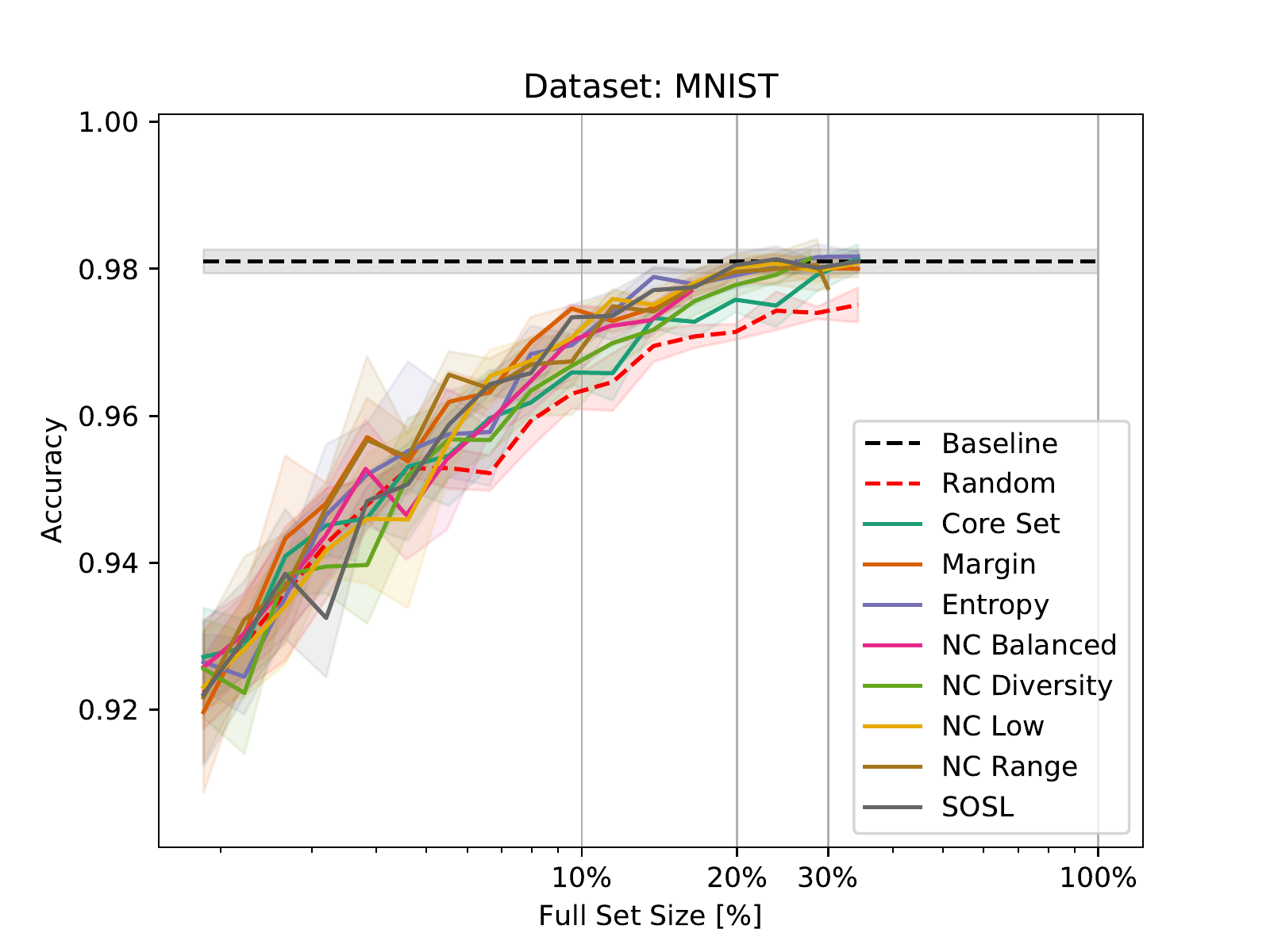} & \hspace{0.0cm}
	     	\includegraphics[clip, trim=0.5cm 0.35cm 1.65cm 1.4cm,
	     		page=1,width=.375\textwidth]{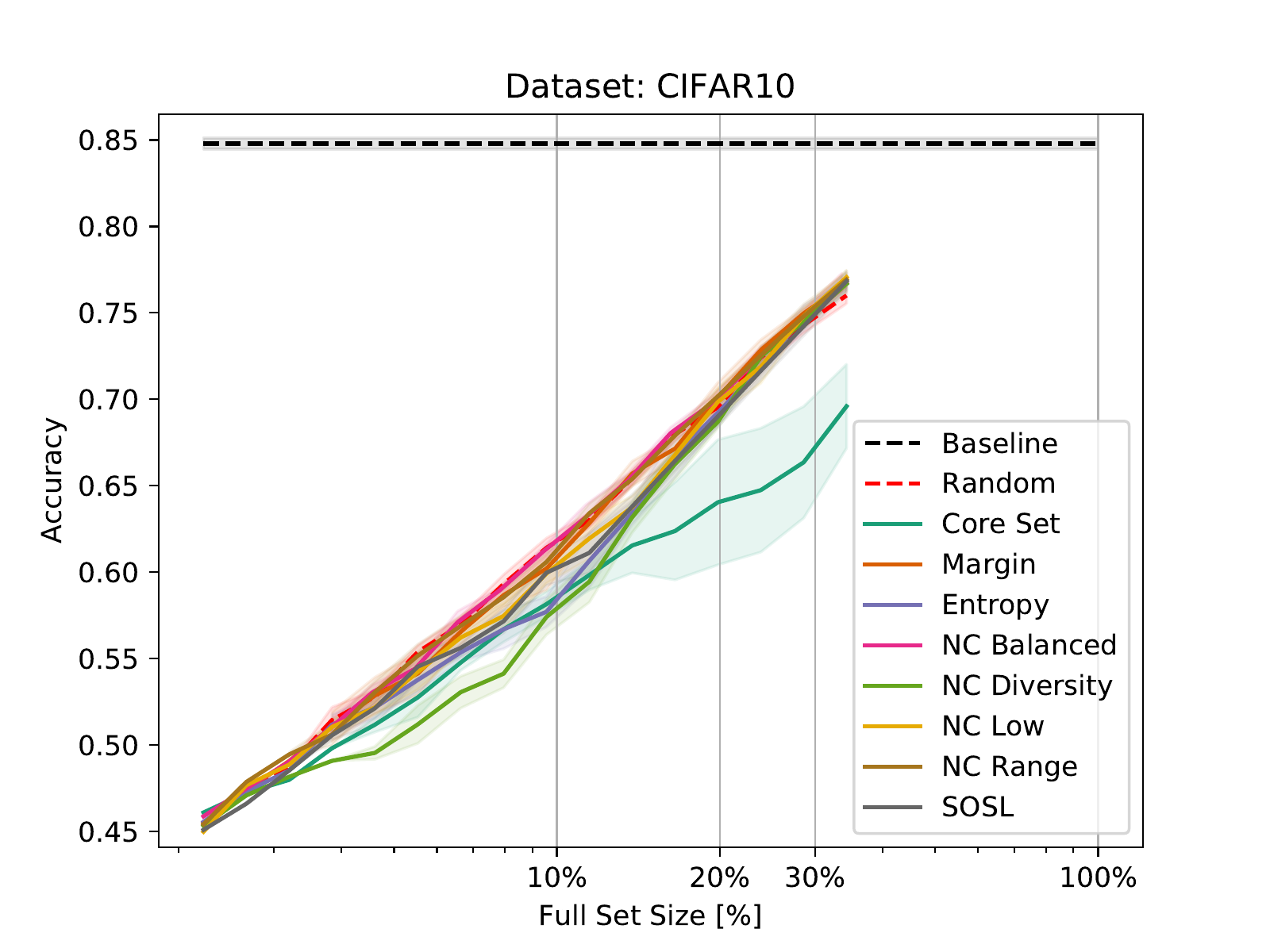}%[.5cm]
       		%includegraphics[page=1,width=.45\textwidth]{images/fMNIST_full_result_std} \\
       		\vspace{-0.2cm}
	   \end{tabular}
	\end{center}
	\caption{Classification accuracy over training set size for all strategies on Fashion-MNIST (top left), TSR (top right), MNIST (bottom left) and CIFAR10 (bottom right). The plotted value is the median of five runs and the shaded area denotes one standard deviation.} %Please note the logarithmic scale of the $x$ axis.}
	\label{fig:generalperformance}
\end{figure}

\subsection{General Performance}
\label{subsec:general}
Before we analyse the robustness of the presented query strategies, we compare their general performance on the datasets presented above. For each dataset we use a distinct plain feed-forward CNN. Only for CIFAR-10 we use an implementation of ResNet50 \cite{resnet}. As we are not aiming to find the best architecture for a certain problem but to identify the most promising samples, we choose the number of layers and channels according to the approximate complexity of the task and select learning rates and batch sizes in commonly used ranges.
% !!! 
% [Details im Anhang, Dropout und co. ??]
% !!!

For all of these experiments, we start with a training set of $100$ samples per class of the particular dataset. We train the CNN for up to $1000$ epochs with an early stopping of $200$. For this purpose we split $10\%$ of the training set into an additional "development set". It is not used for training but to validate classification over the course of the training. This is done to obviate an overfitting-like bias with the use of early stopping. Of course the validation accuracy is then determined on the original test set of the respective dataset, using the best network weights acquired during training according to the development set accuracy.

This network is also the one used to then select new samples to be added to the training set utilizing the query strategies. With each iteration we increase the number of samples in the training set by $20\%$. In all cases we conduct five repetitions per strategy and dataset for statistical significance. To reduce the computational burden, %of this large scale experiment,
we iteratively draw new samples until we have reached approximately a third of the full size of the respective training set.

Figure \ref{fig:generalperformance} illustrates the results of the evaluation of all query strategies. %on three of the six datasets.
% !!! 
% (additional results can be found in the supplemental material).
% !!! 
Nearly all findings show a benefit of Active Learning methods and at least some of the query strategies are either hitting the baseline, or are close to it, around the $30\%$ mark. For CIFAR-10 however, this is not true. None of the methods show any profit for this dataset and are in line with the random sample selection, resulting in a nearly perfectly linear increase in accuracy. This does not come as a surprise, as CIFAR-10 has very diverse representations of its classes and seems to contain no redundant information.
\begin{figure}%[h!]
	\begin{center}
		\begin{tabular}{c@{}c@{}}%{@{}c@{\hspace{.5cm}}c@{}} 
			\hspace{-0.5cm}
       		\includegraphics[clip, trim=0.3cm 0.8cm 0.2cm 0.9cm, page=1,width=.55\textwidth]{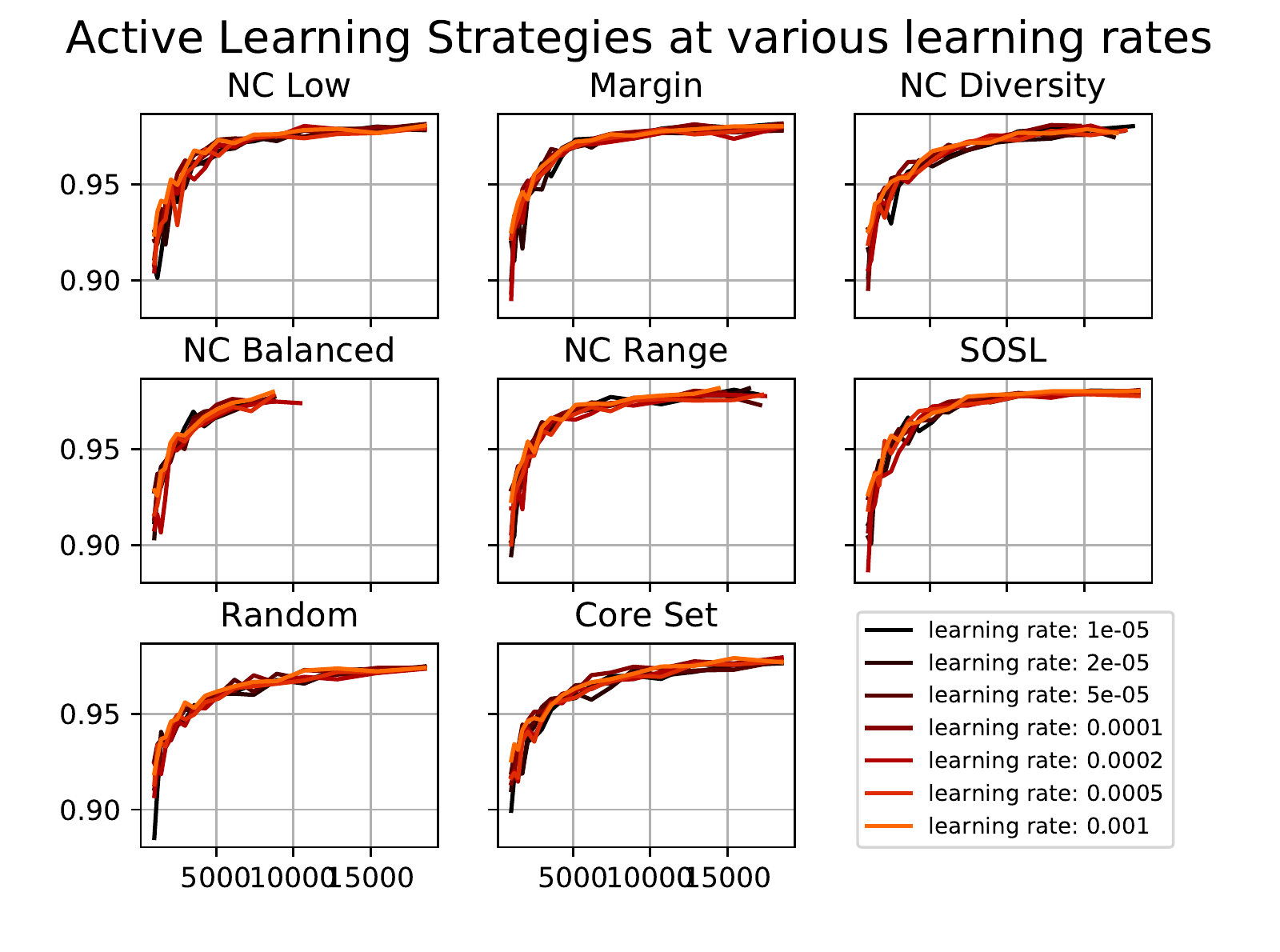} & \hspace{-0.55cm}
       		\includegraphics[clip, trim=0.3cm 0.8cm 0.2cm 0.9cm, page=1,width=.55\textwidth]{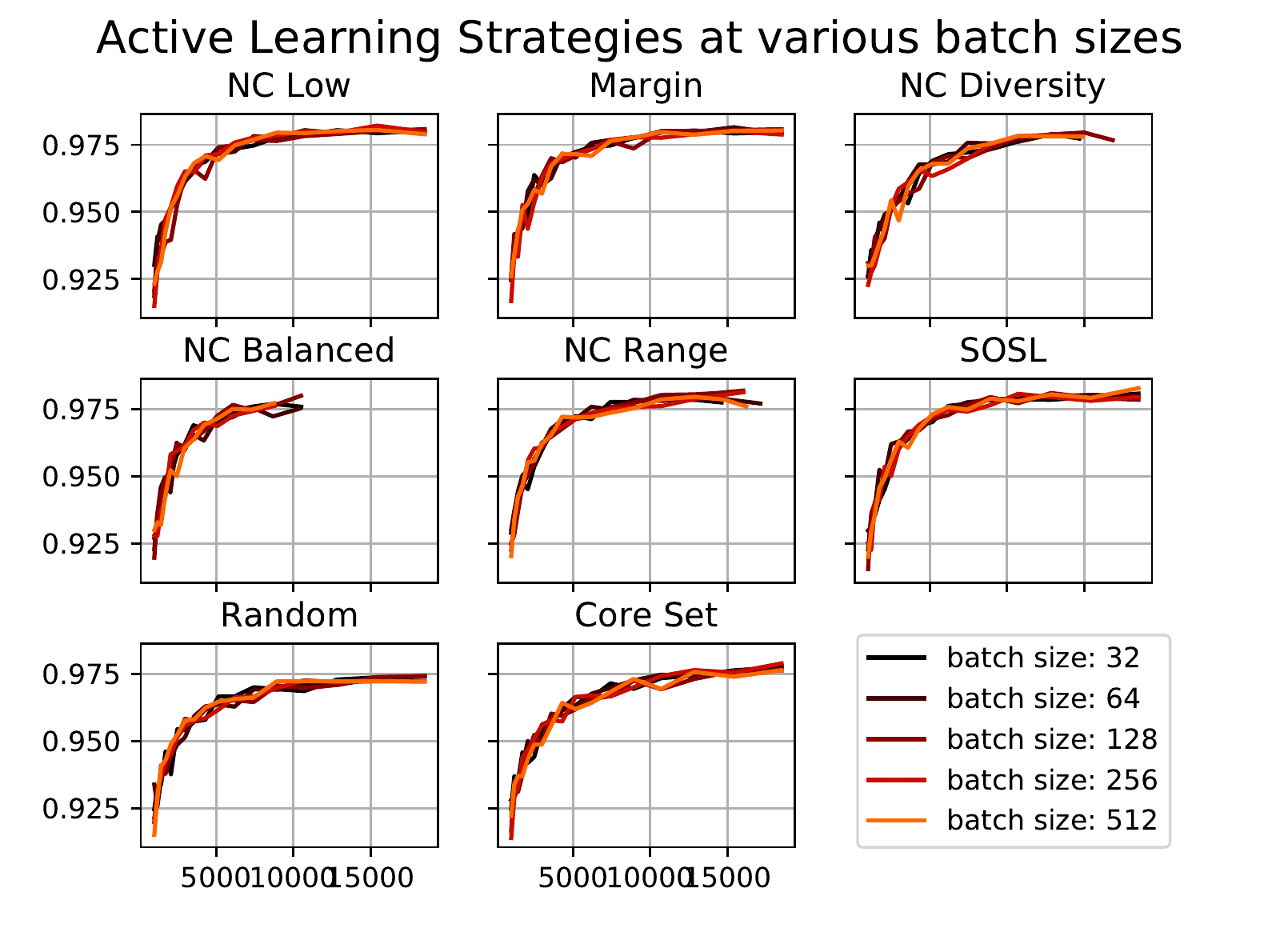} \\
	   \end{tabular}
	\end{center}
	\caption{Changes in learning rate (left) and batch size (right) for various Active Learning strategies on MNIST. The plotted value is the median of five runs.}
	\label{fig:hyperparams}
\end{figure}

\subsection{Changing Hyperparameters and Falsely Labelled Data}
\label{subsec:hyperparams}

As hyperparameter optimisation is very important in fine-tuning the performance of Machine Learning algorithms, we analyse how much changes in these parameters influence the usability of the Active Learning methods shown.\\
Figure \ref{fig:hyperparams} shows the effect of altering the learning rate over two magnitudes and the batch size up to a factor of $16$, for experiments on MNIST. All methods behave very robustly and do not show to be influenced by these alterations.

\begin{figure}%[h!]
	\begin{center}
		\begin{tabular}{c@{}c@{}}	
			\hspace{-0.5cm}
       		\includegraphics[clip, trim=0.3cm 0.8cm 0.2cm 0.9cm, page=1,width=.55\textwidth]{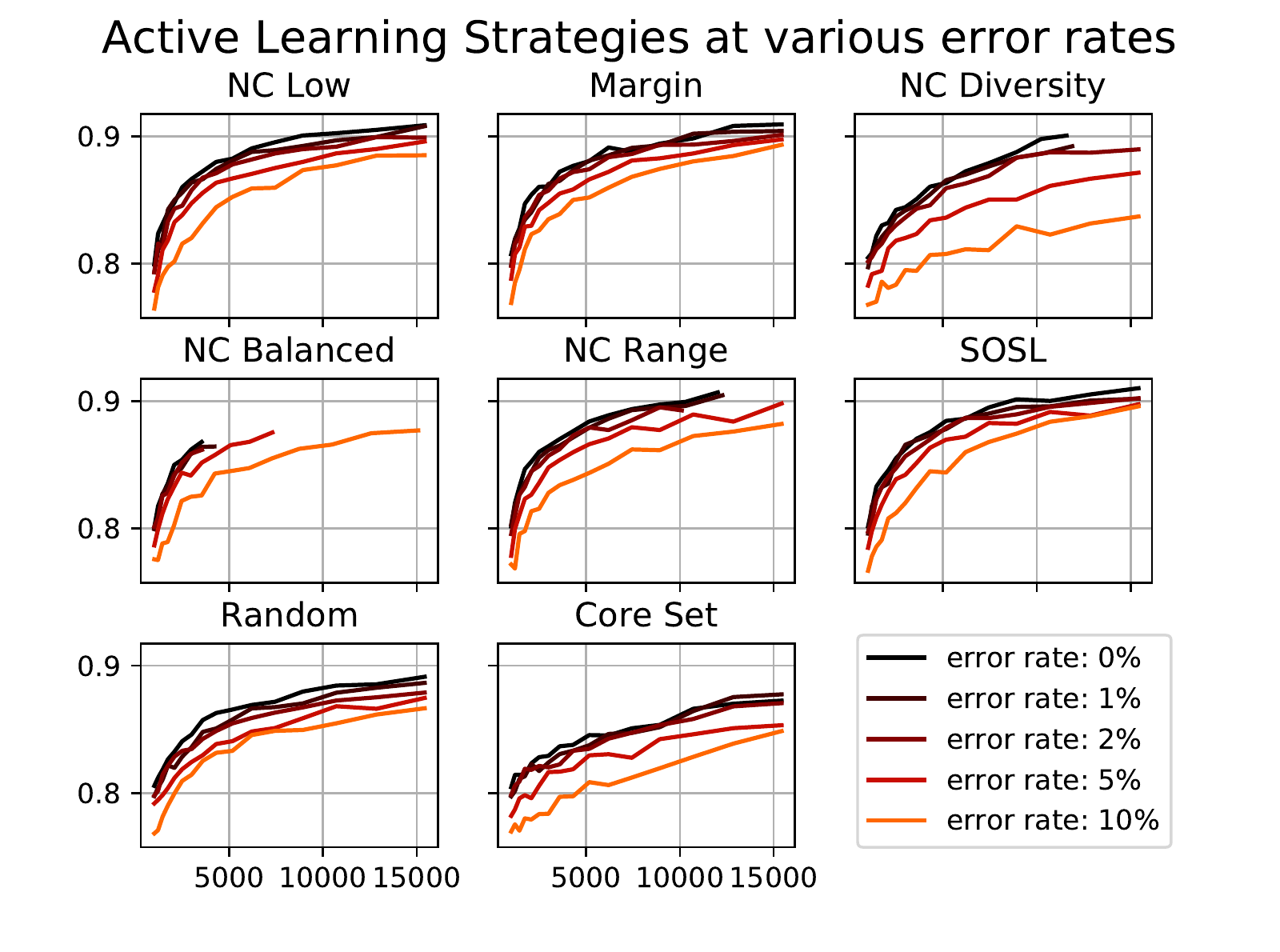} & \hspace{-0.55cm}
       		\includegraphics[clip, trim=0.3cm 0.8cm 0.2cm 0.9cm, page=1,width=.55\textwidth]{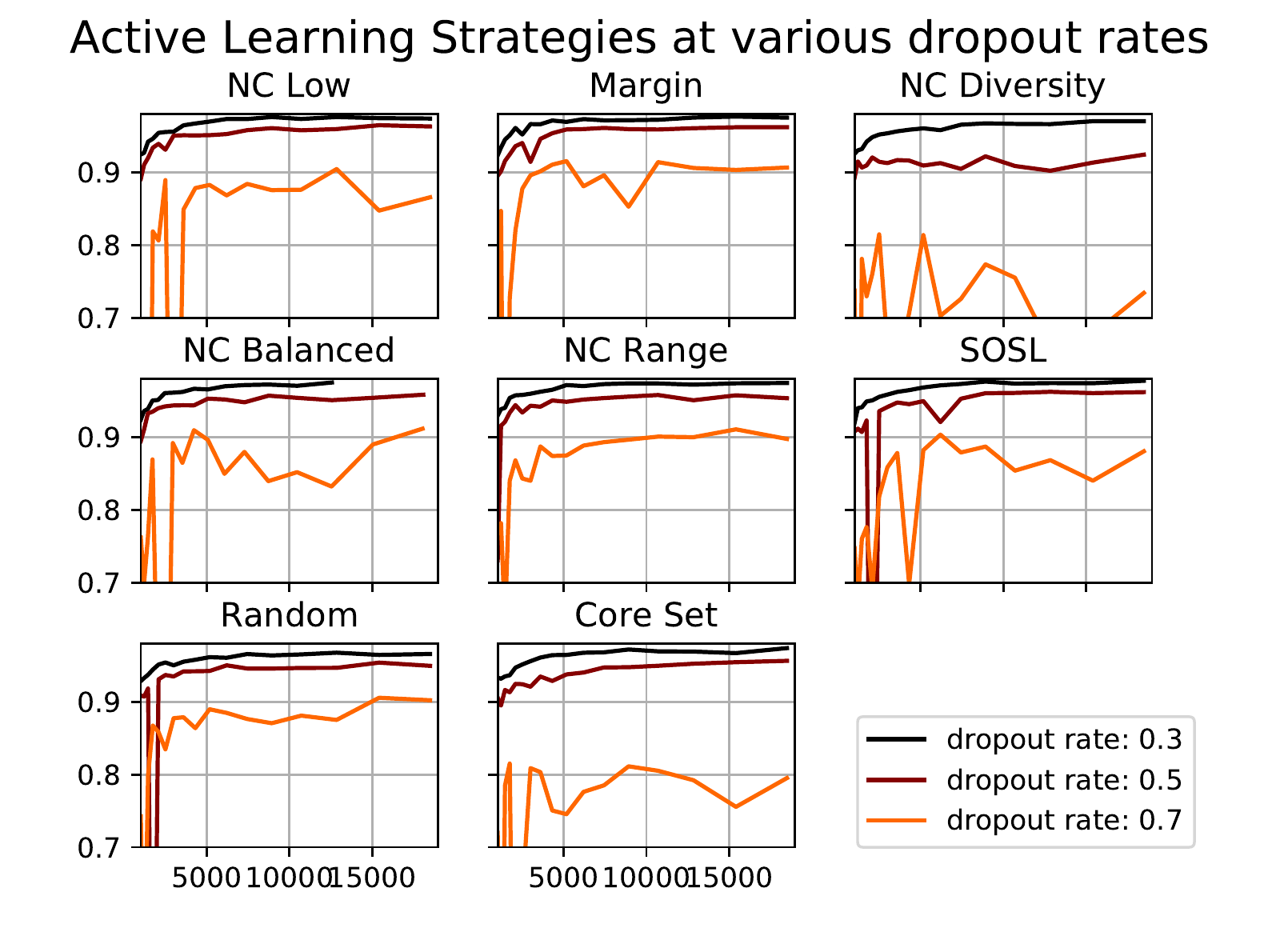} \\
       	\end{tabular}
	\end{center}
	\caption{Active Learning strategy results for various synthetic labelling error rates on Fashion-MNIST (left) and different dropout regularization rates on MNIST (right). The plotted value is the median of five runs.}
	\label{fig:errors}
\end{figure}

Since it can be expected that human annotation, especially in large scale labelling of sensor data, is never perfectly accurate, it is interesting to investigate how this might interfere with the applicability of Active Learning. In Figure \ref{fig:errors} (left) we show results for an experiment where we purposely introduced false labels into the Fashion-MNIST training set. It can clearly be seen, that methods relying on a diversity criterion (NC Diversity, Core Set) suffer the most, since their selection process prevents similar samples from being chosen and therefore it can be harder to correct the negative impact that the selection of a wrongly labelled sample would have. Please note that these strategies also show the highest sensitivity to changes in dropout (cf. Figure \ref{fig:errors} right).

\subsection{Replaceability of Classifiers}
\label{subsec:classifiers}
%min: 20 + 76 + 74 + 1020 + 800 + 400 = 2390
%med: 80 + 438 + 330 + 330 + 220 + 1950 + 1800 + 600 = 5748
%max: 320 + 9248 + 18496 + 36928 + 73856 + 262272 + 32768 + 2560 = 436448
%("Min": $2.4k$ parameters, "Med": $5.8k$, "Max": $436.5k$)

%(C2 [conv. layer with 2 channels], C4, MP (max pooling), C2, MP, $5\times5$C20, FC40 (fully connected with 40 neurons), FC10)
%(C8, C6, C6, C6, MP, C4, MP, C30, FC60, FC10)
%(C32, C32, C64, C64, MP, C128, MP, C128, FC256, FC10)

In the application of Machine Learning, especially in product context, successive refinement of the algorithm is very common. A CNN architecture might be adjusted several times over the course of development or a production process, to optimise the performance or to adapt to changes in the dataset or external restrictions like computational resources. We investigate how the usability of Active Learning might be influenced, if data selection is done by a different network than the one eventually targeted for classification performance. For this purpose we implemented three CNNs, referred to as $Min$, $Med$ and $Max$ in the following, of different capacity to iteratively select samples from Fashion-MNIST with the query strategies as described above. We then perform a cross-training, where every network is trained with the selections of the others and its own. To ensure comparability, we use the same initial dataset of $100$ samples per class for all classifiers and repeat calculations five times.

\begin{figure}[H]
	\begin{center}
		\begin{tabular}{c@{}c@{}c@{}}%{@{}c@{\hspace{.5cm}}c@{}} 
			\hspace{-0.7cm}
	   	 	\hspace{0.15cm}
    	 	\includegraphics[clip, trim=0.3cm 0.15cm 1.6cm 1.4cm, width=0.335\textwidth]{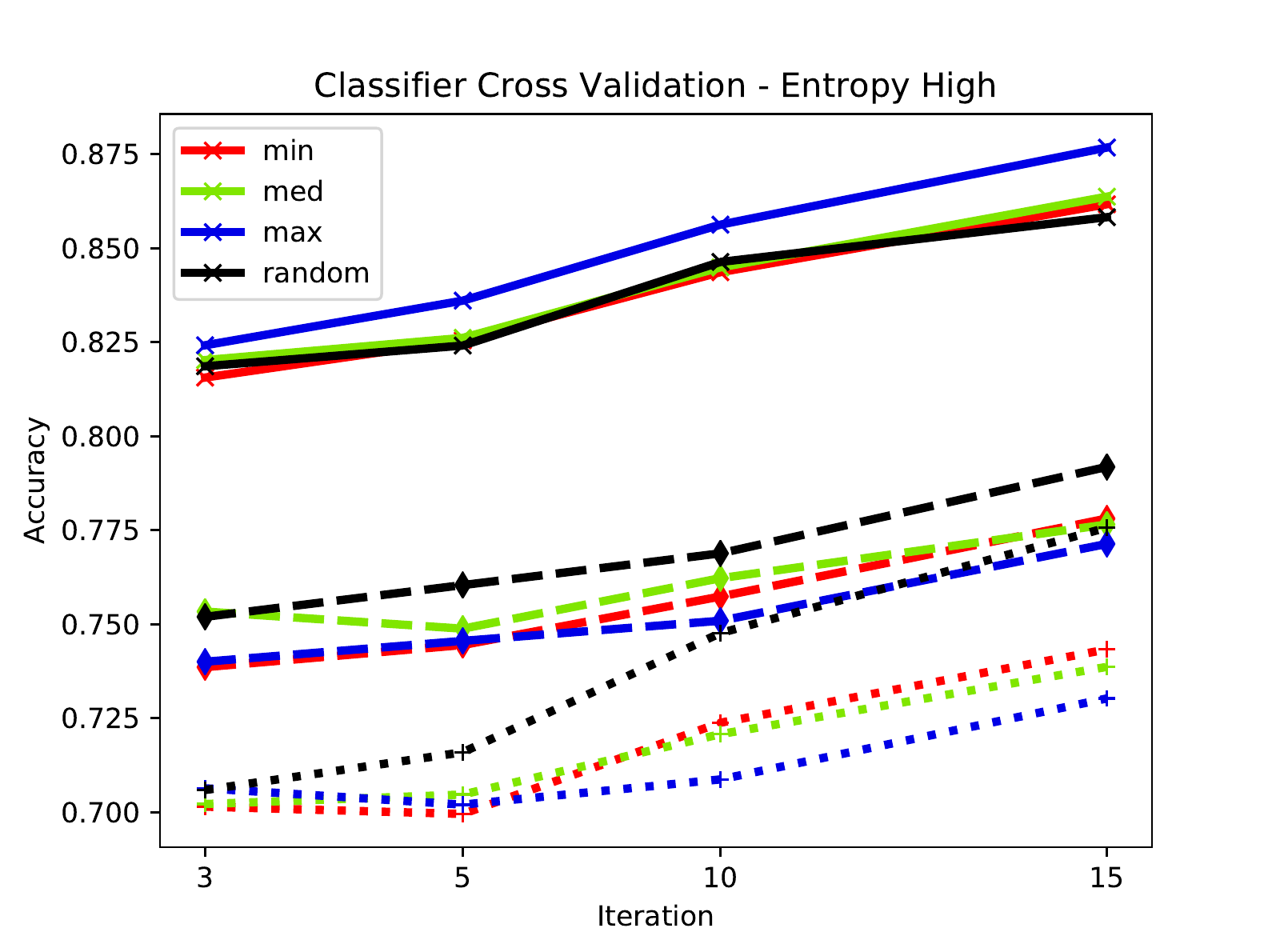} & \hspace{0.25cm}
    	 	\includegraphics[clip, trim=1.8cm 0.15cm 1.6cm 1.4cm, width=0.3\textwidth]{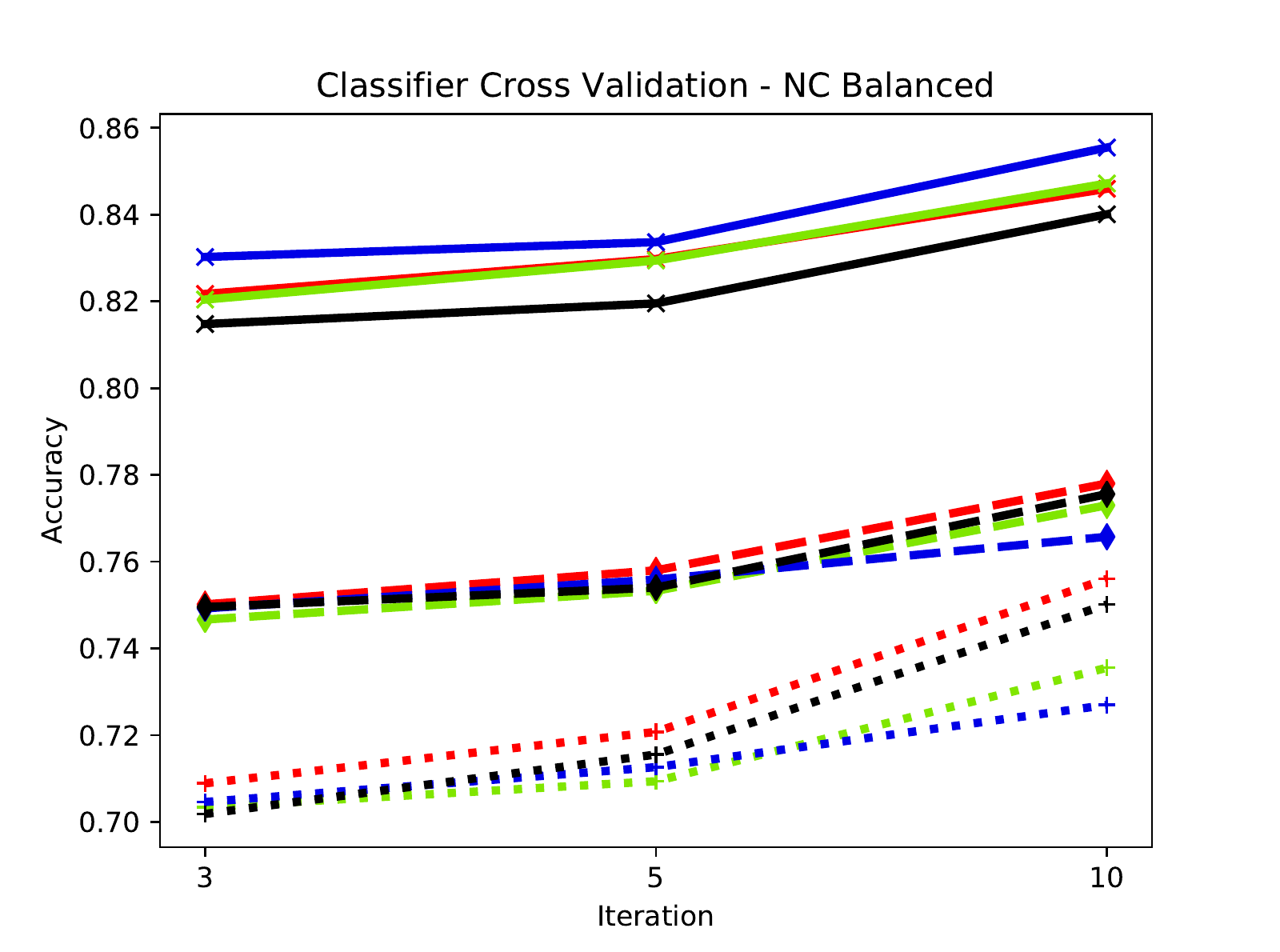} & \hspace{0.25cm}
    	 	\includegraphics[clip, trim=1.8cm 0.15cm 1.6cm 1.4cm, width=0.3\textwidth]{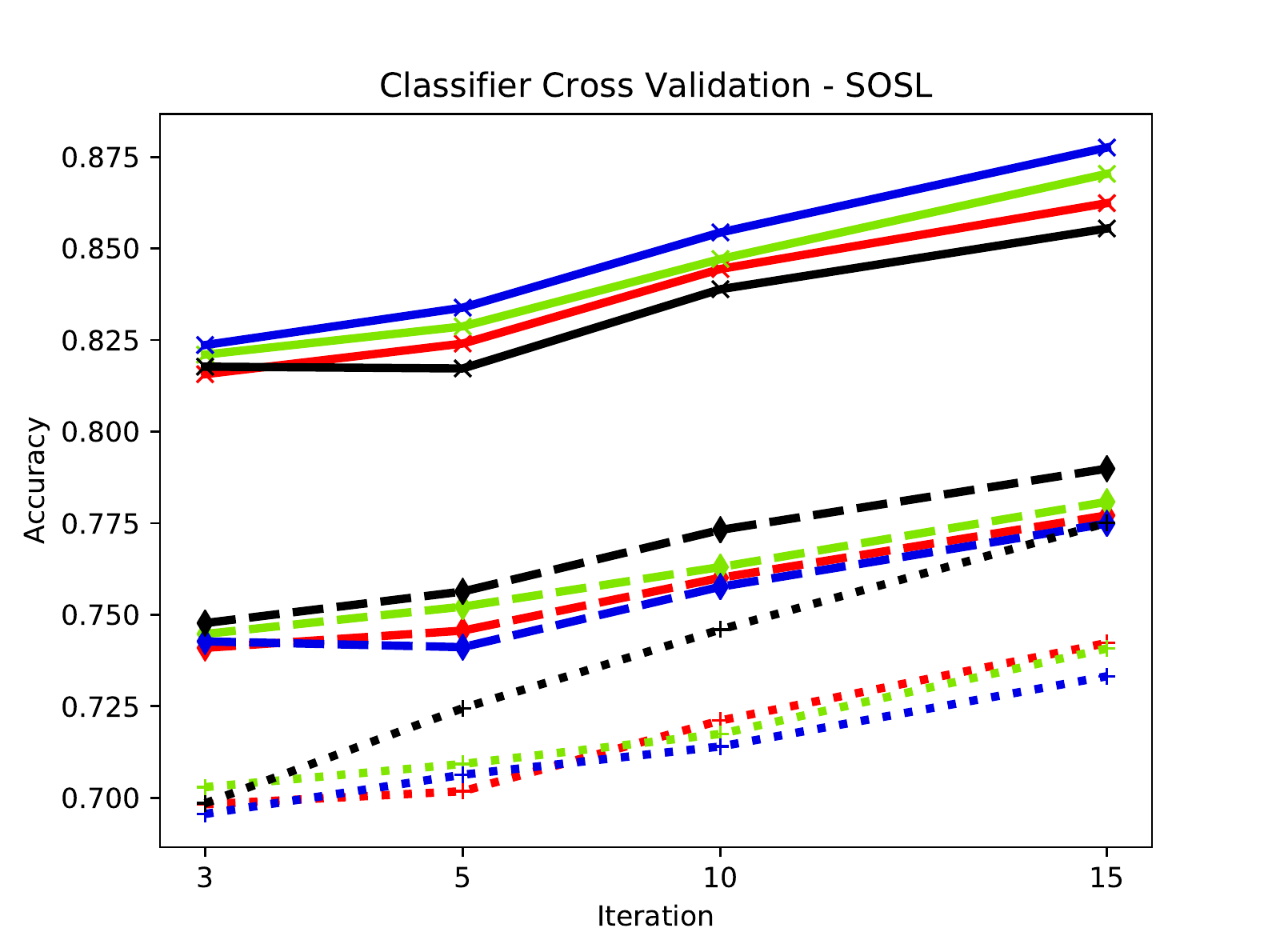}
    	 \end{tabular}
	\end{center}
	\caption{Results (mean of five runs) of cross training of different classifiers, Max (solid line), Med (dashed), Min (dotted), with the sample selection of the other networks and its own (blue, green, red), compared against a random selection (black), for the strategies Entropy High (left),  NC Balanced (mid) and SOSL (right). Evaluations are made after 3, 5, 10 and 15 Iterations of querying new samples, except for NC Balanced which already terminated before the 15th iteration.}
	\label{fig:cross}
\end{figure}

Figure \ref{fig:cross} shows the results for selected strategies.
% !!!
%(more can be found in the supplemental material). 
% !!!
Apart from information about the replaceability of classifiers, these results can show how the classifier capacity itself influences the applicability of Active Learning strategies. For the example of NC Balanced we can note a bias for the own selection performing best with the $Max$ and $Min$ classifier, while the medium-sized one shows indifference. The "weaker" the network gets, the better the performance of the random selection becomes. For the SOSL, this becomes even more clear. While the selection of the $Max$ classifier is still definitely the best for itself, the smaller networks show the best performance with the randomized set. The results with Entropy High are very similar, but the gaps become even more obvious. $Max$ now shows a very clear preference for the own selection compared to any other and the performance of the Active Learning strategy selection on the $Min$ network is now more than three percentage points behind random.

\subsection{Hierarchical Classifiers}
\label{subsec:hierarchical}

To complete our Active Learning robustness study, we examine a neural network structure different from the straightforward CNNs in the preceding sections.

Hierarchical or cascaded classifiers do not use a single label per sample but a whole label tree (cf. \cite{weyers-etal_18}). Consequently, label vectors consist of one of the three following options per class: "1, 0 or not applicable" and each sample belongs to exactly one class per hierarchy level. Furthermore, during the learning phase each class is treated independent of all others. If we have an $n$-class classification problem, $n$ "1-vs-all classifiers" are trained.

\begin{figure}[t]%[H]
	\centering
	%\begin{center}
		% left bottom right top
		\includegraphics[page=1, clip, trim=0.5cm 0.3cm 0.5cm 1.0cm, width=0.55\textwidth]{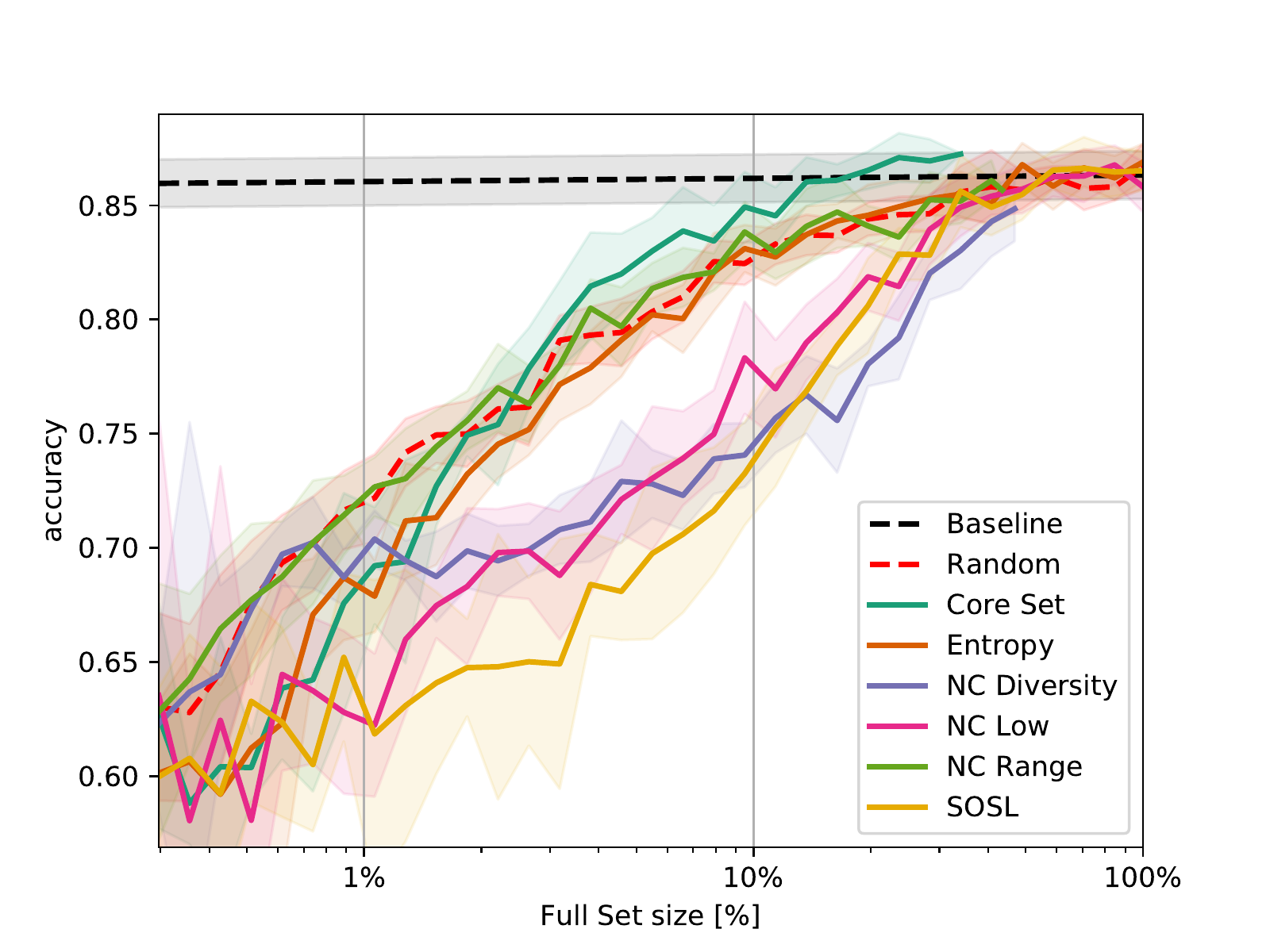} 
	%\end{center}
	\caption{Classification accuracy over training set size for different Active Learning methods on hierarchical classifier for hand gesture classification. The plotted value is the median of ten runs and the shaded area denotes one standard deviation.}
	\label{fig:cascade}
\end{figure}

This renders all Active Learning strategies which rely on quantifying the uncertainty of the logits useless.  All of them (e.g. Naive Certainty, Margin) implicitly rely on the assumption that labels with two possible states are used. As the neurons that belong to classes marked as "not applicable" are not considered during backpropagation (cf. \cite{weyers-etal_18}) they can take arbitrarily high values and thus confuse the mentioned Active Learning methods.
As can be seen in Figure \ref{fig:cascade} this can even result in worse performance than random sampling. However, we can show, that methods, which work in the embeddings space (like the Core Set method), are not effected and thus are also employable for hierarchical neural networks.

\subsubsection{Used Dataset}
\label{subsubsec:dataset}
In all experiments with the hierarchical classifier we use a private dataset that consists of 12 classes which depict different poses of a human hand (e.g. "One finger", "Two fingers", "Fist Thumb Left", etc.). We use a training set of $670\,000$, a development set of $75\,000$ and a test set containing $8\,000$ grey scale images of size $22\times46$.

\begin{figure}[t]%[H]
	\begin{center}
		\includegraphics[width=0.9\textwidth]{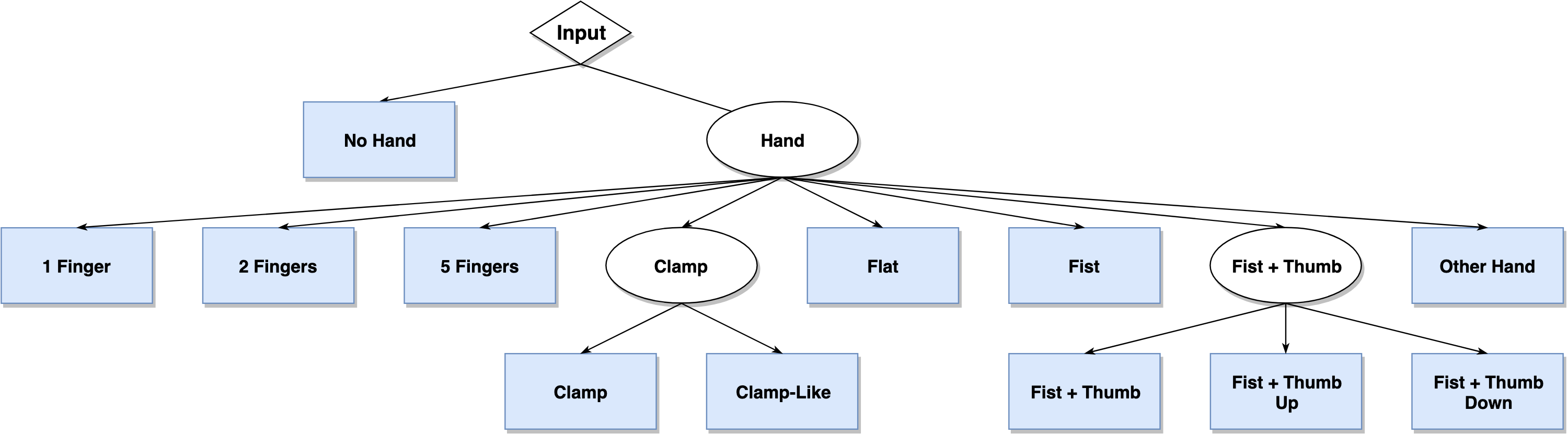}
	\end{center}
	\caption{Hierarchical labels for hand gesture recognition. Blue boxes denote the $12$ classes.}
	\label{fig:gesture_structure}
\end{figure}

As depicted in Figure \ref{fig:gesture_structure}, we use three levels of hierarchy: 1.) "Hand"/"No hand", 2.) Class, 3.) Subclass. A sample of "Fist Thumb Left" e.g. would have the labels "Hand + "Fist Thumb" + "Fist Thumb Left". Especially the neurons of the subclasses often have the label "not applicable" as each subclass belongs to only one class.

%\begin{figure}%[H]
%	\begin{center}
%		\begin{tabular}{c@{}c@{}}
%			\hspace{-0.7cm}
%    	 	\includegraphics[page=1,width=0.55\textwidth]{images/gesture} & \hspace{-0.45cm} 
%    	 	\begin{minipage}{0.485\textwidth}
%    		\vspace{-5cm}
%    	 	\includegraphics[width=\textwidth]{images/gesture_graph} 
%    	 	\end{minipage}
%    	 \end{tabular}
%	\end{center}
%	\caption{Example of a short caption, which should be centered.}
%	\label{fig:cascade}
%\end{figure}

%------------------------------------------------------------------------
\section{Conclusion}

We have presented a study on the robustness of Active Learning. While we show that even plain methods can bring a notable profit in different image classification applications, we emphasise, that prior knowledge about the data and the Machine Learning algorithm in use is essential for successful application. As seen in \ref{subsec:general}, methods that work well on a number of datasets might suddenly fail on a different one and certain data collections might be inherently unsuitable for this kind of active data selection. Although many changes in hyperparameters and erroneous labels might not influence the performance of particular strategies on one hand (cf. \ref{subsec:hyperparams}), classifier changes on the other can by all means (cf. \ref{subsec:classifiers}). Critical alterations in the way a Machine Learning tasks is tackled, like switching from a straightforward to a hierarchical classifier (cf. \ref{subsec:hierarchical}), can turn the all previous findings upside down. \\
These findings underline, that Active Learning can be a helpful tool in data science, but has to be used with knowledge about the targeted utilisation. We aim to continue our endeavours in this field and expand the considerations to segmentation problems and ways to automatically provide assessment on promising combinations of data, Machine Learning algorithms and Active Learning strategies, to avoid possible pitfalls like the ones presented in this work.

%------------------------------------------------------------------------

\label{sect:bib}
\bibliographystyle{plain}
%\bibliographystyle{alpha}
%\bibliographystyle{unsrt}
%\bibliographystyle{abbrv}
% Bibliography
\bibliography{literature}

%------------------------------------------------------------------------------
% Index
%\printindex

%------------------------------------------------------------------------------
\end{document}